\documentclass[10pt,journal,compsoc]{IEEEtran}

\usepackage[bb=boondox]{mathalfa}
\usepackage{hyperref}       
\usepackage{url}            
\usepackage{booktabs}       
\usepackage{nicefrac}       
\usepackage{microtype}      
\usepackage{xcolor}         

\usepackage{amsmath, amssymb, amsfonts, amsthm, mathtools}
\usepackage{algorithm, algorithmicx, algpseudocode}
\usepackage[pdftex]{graphicx}

\usepackage{caption}
\newtheorem{theorem}{Theorem}
\newtheorem*{theorem*}{Theorem}
\newtheorem{lemma}{Lemma}

\newtheorem{corollary}{Corollary}
\newtheorem{definition}{Definition}
\newcommand{\centered}[1]{\overline{#1}}
\newcommand{\doublecentered}[1]{\centered{\centered{#1}}}

\newcommand{\comment}[1]{}
\newcommand{\demo}[1]{\noindent{\em #1}}

\newcommand{\boxenddemo}{$\Box$}

\newcommand{\euc}[1]{\frac{|#1|^2}{2}}
\ifCLASSOPTIONcompsoc
  \usepackage[nocompress]{cite}
\else
  \usepackage{cite}
\fi

\begin{document}

\title{Data eccentricity, asymptotics of Gaussian RBF reproducing kernel Hilbert space, and kernel PCA}

\author{Sergio A.\ Alvarez

\IEEEcompsocitemizethanks{\IEEEcompsocthanksitem Generative AI was not used in carrying out this work 
or in writing this paper,
except possibly as embedded by default in standard search engines. S.\ A.\ Alvarez is with the
Department of Computer Science, Boston College, Chestnut Hill, MA 02467 USA. 
E-mail: alvarez@bc.edu
}

} 

\IEEEtitleabstractindextext{%
\begin{abstract}
We show that, up to isotropic scaling, 
the Gaussian RBF reproducing kernel Hilbert space (RKHS) is asymptotically isometric to Euclidean space in the large bandwidth limit. This strongly suggests that kernel-based constructions reliant on metric properties of the RKHS will 
yield results for Gaussian RBF kernels that similarly approach those of linear kernels for large bandwidths. 
The asymptotic behavior of Gaussian CKA can be understood in this light. We further consider kernel PCA, 
showing that Gaussian RBF eigenvalues, 
eigenprojections, and principal components all converge to those of classical (linear) PCA as 
bandwidth $\sigma \rightarrow \infty$. For a given data representation, both the RKHS feature embeddings and the orthogonal PCA eigenframes of the two kernel types differ asymptotically by a geometric similarity transformation, 
up to a residual of size 
$O \left (\frac{\rho}{\sigma} \right )^2$, 
where $\rho$ is a measure of geometric eccentricity of the representation, equal to the ratio of maximum to
median pairwise distance between data examples. 
Experiments over a diverse collection of data sets demonstrate that $\rho$ provides a simple and reliable predictor of dataset-specific convergence behavior in the top principal directions. 
\end{abstract}

}

\maketitle

\IEEEdisplaynontitleabstractindextext

%



\IEEEraisesectionheading{\section{Introduction}}

If $k: \mathbb R^d \times \mathbb R^d \rightarrow \mathbb R$ is a positive semi-definite symmetric kernel function, the Moore-Aronszajn theorem~\cite{Aronszajn1950} guarantees the existence of a feature map $\Phi: \mathbb R^d \rightarrow \mathcal H$ from the original attribute space $\mathbb R^d$ into a typically much higher-dimensional reproducing kernel Hilbert space (RKHS) $\mathcal H$,
with the property that, for any data examples $x_i$ and $x_j$, the value $k(x_i,x_j)$ is the inner product 
$\langle \Phi(x_i), \Phi(x_j) \rangle_{\mathcal H}$ of the feature vectors of those data examples in $\mathcal H$.
Any computation that can be expressed in terms of pairwise inner products between feature vectors can therefore be carried out efficiently by the ``kernel trick'' (e.g.,~\cite{scholkopf2000kernelTrickForDistances}), replacing complex inner product computations in the high-dimensional RKHS by equivalent, but much simpler, evaluation of the kernel function in the original low-dimensional attribute space.  

The choice of kernel function is important, as it determines the RKHS feature representation. Gaussian Radial Basis Function (RBF) kernels are a popular kernel type corresponding to an infinite-dimensional RKHS, and satisfy a powerful universal approximation property~\cite{micchelli2006universal},~\cite{KernelUniversality2011}. The latter is particularly important
in light of the ``representer theorem''~\cite{kimeldorf1971some},~\cite{wahba1999support},~\cite{scholkopf2001generalized} that states that kernel-based predictive models that minimize an empirical risk function are in the span of the functions $k(\cdot, y)$ for fixed $y$: kernels that satisfy a universal approximation property yield hypothesis spaces that approximate any target function arbitrarily closely. 

Gaussian RBF kernels have been used with success in a variety of machine learning methods and applications, including support vector machines (SVM)~\cite{cortes1995support,scholkopf1997comparing},~\cite{varewyck2010practical},~\cite{ring2016approximation},~\cite{kang2009chip},~\cite{xiao2014parameter},~\cite{tang2009efficient};
kernel ridge regression~\cite{avron2017random},~\cite{chavez2020scalable},~\cite{wu2020predicting},~\cite{ding2022kernel},~\cite{pmlr-v30-Zhang13},~\cite{zhang2015kernel},~\cite{exterkate2013model},~\cite{burnaev2016conformalized};
centered kernel alignment (CKA)~\cite{CortesEtAl2010, CortesCKA2012},~\cite{KornblithCKA2019, KornblithWideDeepCKAICLR2021},~\cite{CKAinNLP2020},~\cite{CKAforRNNSimilarity2019},~\cite{DrugSideEffects2019},~\cite{BrainActivityPatterns2017},~\cite{CKAforWatermarkEntanglement2021};
canonical correlation analysis (CCA)~\cite{bilenko2016pyrcca},~\cite{gorecki2020independence},~\cite{chen2019graph},~\cite{fukumizu2007statistical},~\cite{feng2008real},~\cite{zheng2006facial};
and 
kernel principal component analysis (kernel PCA)~\cite{ScholkopfEtAlKernelPCA1997},~\cite{ScholkopfEtAlKernelPCA1998},~\cite{jorgensen2011model},~\cite{pmlr-v51-ghashami16},~\cite{AlaizConvexOptimizationKernelPCA2018},~\cite{HOFFMANN2007863},~\cite{abba2020emerging},~\cite{MikaScholkopfKernelPCADenoising1998},~\cite{rosipal2001kernel}. 

\subsubsection*{Contribution of the present paper}
We prove that, up to an isotropic scale factor, the 
RKHS associated with a Gaussian RBF kernel is geometrically identical to the linear kernel RKHS in the limit as the Gaussian bandwidth hyperparameter approaches infinity. We show that the metric residual is of order $O \left ( \rho/\sigma \right )^2$, where $\rho$ reflects the geometric eccentricity of the data set as measured by the ratio of maximum to median pairwise distance between data examples (Theorem~\ref{thm:GaussianPCAResidualDetails}). 

In light of our results in Theorem~\ref{thm:GaussianPCAResidualDetails}, constructions based on the metric structure 
of the RKHS can be expected to yield behaviors for Gaussian RBF kernels that approach those of linear kernels for bandwidths $\sigma \gg \rho$. 
The asymptotics of Gaussian RBF CKA~\cite{AlvarezGaussianCKA2022} comprise one example of this principle. 
We provide an additional illustration in the present paper, by using spectral perturbation theory~\cite{KatoPerturbationTheoryBook1995}
to show that Gaussian kernel PCA is asymptotically equivalent to classical linear PCA as bandwidth $\sigma \rightarrow \infty$ (Theorem~\ref{thm:GaussPCASpectralLimit}).  

Experiments over a range of data sets demonstrate that eccentricity, $\rho$, reliably gauges dataset-specific convergence of the Gaussian RKHS metric structure and kernel PCA eigenvalues and eigenprojections in the top principal directions. 

\subsubsection*{Related work}
Both large- and small-bandwidth limits for support-vector machines (SVM) based on Gaussian kernels are discussed
from the perspective of the dual optimization formulation of SVM in~\cite{KeerthiLinGaussianSVMLimits2003}. In~\cite{barthelme2021spectral}, the authors obtain spectral asymptotics results for a wide class of kernels that includes
Gaussian RBF kernels. Large-bandwidth Gaussian asymptotics for Gaussian process regression are discussed 
in~\cite{barthelme2022gaussian}. All of these works focus on uncentered kernels, however.
 
Work involving the large-bandwidth limit for centered kernels as used in the present paper 
includes~\cite{amblard2022mesures}, which discusses the related Hilbert-Schmidt Independence Criterion (HSIC)
for a wide class of kernels, 
and~\cite{AlvarezGaussianCKA2022}, which deals with Gaussian RBF centered kernel alignment (CKA).
The data eccentricity measure used in the present paper was introduced in~\cite{AlvarezGaussianCKA2022};
that paper showed that eccentricity is relevant to the large-bandwidth convergence of Gaussian RBF CKA, 
but did not uncover the root of this behavior in the geometry of the Gaussian RBF RKHS, as we do here.

We are not aware of prior work that addresses the large-bandwidth asymptotics of either the Gaussian RBF RKHS 
metric space structure or of Gaussian RBF kernel principal components analysis (PCA), which are the main issues 
considered in the present paper.

\section{Background}

\subsection{Kernel PCA}

\begin{figure}[h]
	\begin{center}
		\includegraphics [width=0.7\columnwidth, clip=true, trim=0mm 5mm 0mm 10mm]
		{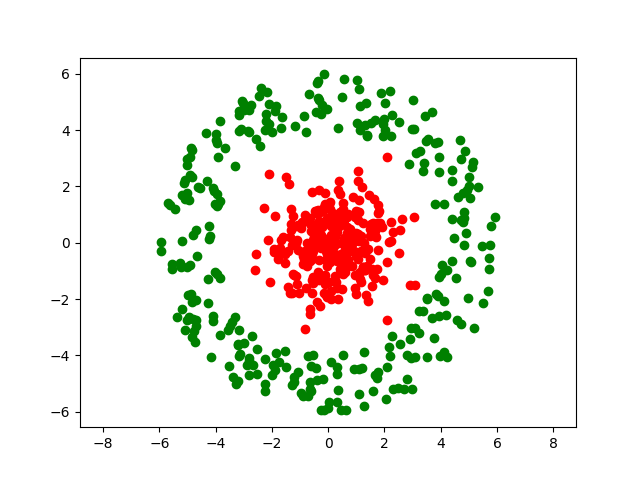}
	\end{center}
	\caption{Nonlinearly co-varying attributes with null covariance.}
	\label{fig:ringedGaussian}
\end{figure}

Principal Components Analysis (PCA) computes the directions of maximum variance of a set of data examples, $X = \{x_i,\ i=1...N\} \subset \mathbb R^d$, together with the variances in those directions, as 
the eigendirections and eigenvalues of the sample covariance matrix, $\frac{1}{N-1}X^T X$ (we assume that $X$ contains one data example per row, and that the data are centered in attribute space). The eigendirections comprise an orthogonal reference frame for $\mathbb R^d$ that aligns with the axes of linear co-variation of the data attributes. 
As is well known, covariance does not capture nonlinear co-variation; for example, in Fig.~\ref{fig:ringedGaussian},
the sample covariance matrix is the identity, incorrectly suggesting an absence of co-variation.

Kernel PCA~\cite{ScholkopfEtAlKernelPCA1997, ScholkopfEtAlKernelPCA1998} carries out PCA on a high-dimensional feature representation $\Phi(X)$ of $X$ in an RKHS, $\mathcal{H}$, defined implicitly by means of a positive semi-definite kernel function, 
$k(x,y) = \langle \Phi(x), \Phi(y) \rangle_{\mathcal{H}}$, as in the Moore-Aronszajn theorem. 
Classical PCA is the special case of a linear kernel: $k(x,y) = x \cdot y$, where $\cdot$ denotes the standard inner product in Euclidean space. 

\subsubsection{Kernel PCA procedure}
\label{section:KernelPCA}
We describe the kernel PCA computation, following~\cite{ScholkopfEtAlKernelPCA1997, ScholkopfEtAlKernelPCA1998}. 
Assume that the original data comprise an $N \times d$ matrix, $X$, in which the $i$-th row, $x_i$, contains the 
raw feature vector (in $\mathbb R^d$) of the $i$-th data example. 
Let $k(x,y)$ be a positive semi-definite kernel function, with associated RKHS feature map $\Phi$, so that $k(x,y)$ is 
the inner product $\langle \Phi(x), \Phi(y) \rangle_{\mathcal H}$ in the RKHS $\mathcal H$ for all $x, y$ in the original attribute space $\mathbb R^d$. 

\noindent
Kernel PCA proceeds as follows~\cite{ScholkopfEtAlKernelPCA1997, ScholkopfEtAlKernelPCA1998}.
\begin{enumerate}

\item Compute the Gram matrix, $K = (k(x_i,x_j))_{i,j}$, which is of size $N \times N$ independently of $d$.
Perform mean-centering on $K$ (\!\cite{ScholkopfEtAlKernelPCA1998}, appendix B). That is, replace $K$ by the
double-centered matrix $\doublecentered{K} = HKH$ as described below in section~\ref{section:isometryDoubleCentering}.

\item 
\label{item:KernelPCAEigendecomposition} 
Compute the eigenvalues, $\lambda_i$, and eigenvectors, $\alpha^i$, of $K$. Normalize $\alpha^i$ so that 
$\lambda_i \alpha^i \cdot \alpha^i = 1$. The eigenvectors, $V^k$, of the RKHS feature vectors 
$\{\Phi(x_i)\}_{i=1}^N$ 
will be bilinear combinations $V^k = \Phi(X)^T \alpha^k$ of the feature vectors $\Phi(x_i)$
and eigenvectors $\alpha^k$ of $K$. Here, $\Phi(X)$ has $i$-th row $\Phi(x_i)$.

\item
\label{item:KernelPCAProjection} The kernel principal components of each $x$ are the projections, 
$\langle \Phi(x), V^k \rangle_{\mathcal H} = \sum_{j=1}^N \alpha^k_j K(x_j, x)$ 
of the feature vector $\Phi(x)$ onto the eigenvectors, $V^k$.
\end{enumerate}

\subsection{The Gaussian kernel and Gaussian CKA}

We focus on the Gaussian kernel of bandwidth $\sigma$. The Gaussian Gram matrix $K_{G(\sigma)}$ for a data matrix $X$ is as in Eq.~\ref{eq:Gaussian_Gram_matrix}; here, $x_i$ is the $i$-th row of $X$ and $d_X$ is the median distance between rows of $X$~\cite{KornblithCKA2019, AlvarezGaussianCKA2022}. Using $d_X \sigma$ instead of $\sigma$ in the exponent 
of Eq.~\ref{eq:Gaussian_Gram_matrix} makes $\sigma$ a standardized quantity.
\begin{align}
K_{G(\sigma)} &= \left ( e^{-\frac{|x_i-x_j|^2}{2 d^2_X \sigma^2}} \right )_{i,j}
\label{eq:Gaussian_Gram_matrix}
\end{align}
Two closely related kernels also enter into our analysis: the linear kernel and Euclidean (pseudo) kernel,
as in Eq.~\ref{eq:linearEuclideanKernels}.
\begin{subequations}
\begin{align}
L(x_1, x_2) &= x_1 \cdot x_2 \label{eq:kernelsA}\\
E(x_1, x_2) &= -\frac{\left \vert x_1 - x_2 \right \vert^2}{2} \label{eq:kernelsB}
\end{align}
\label{eq:linearEuclideanKernels}
\end{subequations}
Centered kernel analysis (CKA) is a kernel-based similarity metric between data representations~\cite{KornblithCKA2019}. 
It was shown in~\cite{AlvarezGaussianCKA2022} that Gaussian CKA behaves asymptotically linearly for large bandwidths,
and that data eccentricity, measured by the ratio of representation diameter to median pairwise distance 
(Eq.~\ref{eq:rho_single_Gaussian_kernel}), approximates 
the bandwidth at which CKA begins exhibiting near-linear behavior.
See Theorem~\ref{thm:residualDependsOnSigmaVsRho}.
\begin{theorem}
(From Theorems 1, 2 in~\cite{AlvarezGaussianCKA2022}; Eq.~\ref{eq:rho_single_Gaussian_kernel} relies on the assumption that $K$ and $L$ use the same data representation, $X$.)\\
Let $K_{G(\sigma)}$ denote the Gram matrix of representation $X$ corresponding to a Gaussian RBF kernel with bandwidth $d_X \sigma$, where $d_X$ is the median Euclidean distance between data examples in representation $X$, and let $L$ be the Gram matrix of representation $X$ corresponding to some positive semi-definite kernel.
Then: 
\begin{align}
\text{CKA}(K_{G(\sigma)},L) = \text{CKA}(K_{\text{lin}},L) + O \left ( \frac{1}{\sigma^2} \right ) \text{ as } \sigma \rightarrow \infty
\label{eq:GaussianCKAResidual}
\end{align}
The residual in Eq.~\ref{eq:GaussianCKAResidual} is bounded by $C \left ( \frac{\rho}{\sigma} \right )^2$ 
when $\frac{\rho}{\sigma} \le 2$, where $C$ is a finite constant that does not depend on the representation $X$, and
$\rho$ is the representation eccentricity defined in Eq.~\ref{eq:rho_single_Gaussian_kernel}.
\begin{equation}
\rho = \frac{\text{diam}(X)}{d_X}
\label{eq:rho_single_Gaussian_kernel}
\end{equation}
\label{thm:residualDependsOnSigmaVsRho}
\end{theorem}

\section{Results}

We show in Theorem~\ref{thm:GaussianPCAResidualDetails} that, like the Gaussian CKA similarity metric as described in Theorem~\ref{thm:residualDependsOnSigmaVsRho}, the Gaussian kernel RKHS feature representation is asymptotically geometrically identical for large bandwidths to an isotropic scaling of the standard linear RKHS representation,
and that data eccentricity $\rho$ (Eq.~\ref{eq:rho_single_Gaussian_kernel}) estimates the convergence threshold. 
We do this through a direct metric analysis of the respective RKHS feature embeddings via their Gram matrices. 

We then show in Theorem~\ref{thm:GaussPCASpectralLimit} that the asymptotic geometry of the Gaussian RKHS
carries over to Gaussian kernel PCA.

\subsection{Isometric property of double-centering}
\label{section:isometryDoubleCentering}

A positive semi-definite Gram matrix captures the metric structure of the corresponding Moore-Aronszajn RKHS feature embedding. Indeed, if $K$ is such a Gram matrix, the squared distance between embedded examples $\Phi(x_i), \Phi(x_j)$ in the RKHS can be written in terms of inner products:
\begin{equation}
\aligned
\Vert \Phi(x_i) - \Phi(x_j) \Vert^2_{\mathcal H} &= K_{i,i} + K_{j,j} - 2K_{i,j}
\endaligned
\label{eq:metricFromGramMatrix}
\end{equation}

The double-centered version of an $N \times N$ Gram matrix $K$ is $\doublecentered{K} = HKH$; here, 
$H = I - \frac{1}{N} \mathbb{1} \mathbb{1}^T$, where $I \in \mathbb{R}^{N \times N}$ is the identity matrix 
and $\mathbb 1 \mathbb{1}^T \in \mathbb{R}^{N \times N}$ is a matrix of ones.
As noted in~\cite{CortesEtAl2010}, double-centering a Gram matrix, $K$, corresponds to mean-centering the feature vectors, in the sense that $\doublecentered{K}_{i,j} = \langle \Phi(x_i) - \centered{\Phi},\ \Phi(x_j) - \centered{\Phi} \rangle_{\mathcal H}$, where $\centered{\Phi}$ is the mean feature vector. It follows, in particular, that $\doublecentered{K}$ is positive semi-definite. 
Given this relationship between $\doublecentered{K}$ and $K$, it is not surprising that pairwise distances between data examples in the feature embedding associated with $\doublecentered{K}$ are the same as in the embedding associated with $K$. For completeness, we verify this property directly in Lemma~\ref{lemma:doubleCenteringPreservesFeatureDistances}.

\begin{lemma}
For any Gram matrix, $K$, the feature embedding associated with the double-centered Gram matrix $\doublecentered{K} = HKH$ is isometric to the feature embedding associated with $K$ itself.
\label{lemma:doubleCenteringPreservesFeatureDistances}
\end{lemma}
\demo{Proof of Lemma~\ref{lemma:doubleCenteringPreservesFeatureDistances}:}
Consider any pair of examples; denote their feature vectors in the RKHS of $K$ by $x_n$ and $x_m$.
The squared distance between these examples in the feature space is:
\begin{equation}
\Vert x_n - x_m \Vert^2_K
=
K_{n,n} + K_{m,m} - 2 K_{n,m}
\label{eq:distSquaredRKHS}
\end{equation}
The analogous expression for $HKH$ instead of $K$ describes the squared distance in the feature space of $\doublecentered{K}$.
In order to compute this latter squared distance, we use the following detailed description of the entries of 
the product $HKH$:
\begin{equation}
\doublecentered{K}_{i,j}
=
K_{i,j} - \frac{1}{N} \sum_{\beta=1}^N K_{i,\beta} - \frac{1}{N} \sum_{\alpha=1}^N K_{\alpha,j} 
+ \frac{1}{N^2} \sum_{\alpha=1}^N \sum_{\beta=1}^N K_{\alpha,\beta}
\label{eq:microDetailsDoubleCenteredK}
\end{equation}
Using Eq.~\ref{eq:microDetailsDoubleCenteredK}, the $HKH$ analog of Eq.~\ref{eq:distSquaredRKHS} now yields:
\small
$$
\aligned
&\Vert x_n - x_m \Vert^2_{\doublecentered{K}}
=
\doublecentered{K}_{n,n} + \doublecentered{K}_{m,m} - 2 \doublecentered{K}_{n,m}\\
&= 
K_{n,n} - \frac{1}{N} \sum_{\beta=1}^N K_{n,\beta} - \frac{1}{N} \sum_{\alpha=1}^N K_{\alpha,n} + \frac{1}{N^2} \sum_{\alpha=1}^N \sum_{\beta=1}^N K_{\alpha,\beta}\\
&+
K_{m,m} - \frac{1}{N} \sum_{\beta=1}^N K_{m,\beta} - \frac{1}{N} \sum_{\alpha=1}^N K_{\alpha,m} + \frac{1}{N^2} \sum_{\alpha=1}^N \sum_{\beta=1}^N K_{\alpha,\beta}\\
&-2 \left [
K_{n,m} - \frac{1}{N} \sum_{\beta=1}^N K_{n,\beta} - \frac{1}{N} \sum_{\alpha=1}^N K_{\alpha,m} + \frac{1}{N^2} \sum_{\alpha=1}^N \sum_{\beta=1}^N K_{\alpha,\beta}
\right ]
\endaligned
$$
\normalsize
The double sums at far right cancel exactly across the three lines. By symmetry of $K$, the two single sums after $K_{n,n}$ in the top of the three lines are equal; their sum cancels the corresponding term on the bottom line. Likewise, the two single sums after $K_{m,m}$ are equal and cancel the corresponding term on the bottom line. 
The only terms remaining on the right-hand side are precisely those on the right-hand side of Eq.~\ref{eq:distSquaredRKHS}. 
This concludes the proof.
\boxenddemo
\paragraph*{Note} The conclusion of Lemma~\ref{lemma:doubleCenteringPreservesFeatureDistances} fails if only one of column-centering or row-centering is used; the singly centered matrices may not be symmetric (e.g., $K = [\begin{smallmatrix}3 &1\\1 &5\end{smallmatrix}]$).

\subsection{Isometry of the Euclidean and linear RKHS}
\label{section:EuclideanGramMatricesAreLinear}

\begin{lemma}
For a given data set, the double-centered linear Gram matrix is identical to the double-centered Euclidean Gram matrix:
\begin{equation}
\doublecentered{K}_E = \doublecentered{K}_{\text{lin}},
\end{equation}
where the Euclidean (pseudo) kernel is $E(x,y) = -\frac{|x-y|^2}{2}$. 
\label{lemma:doubleCenteredEuclideanGramsAreLinear}
\end{lemma}
\demo{Proof:}
We argue as in the first half of the proof of Lemma $2$ in~\cite{AlvarezGaussianCKA2022},
but using double-centering instead of only column-centering.
Applying Eq.~\ref{eq:microDetailsDoubleCenteredK} to the Euclidean kernel, we have:
\footnotesize
$$
\aligned
&{\doublecentered{K}_E}_{i,j}
=
-\euc{x_i-x_j}\\
&+ \frac{1}{N} \sum_{\beta=1}^N \euc{x_i-x_{\beta}} 
+ \frac{1}{N} \sum_{\alpha=1}^N \euc{x_{\alpha}-x_j}
- \frac{1}{N^2} \sum_{\alpha=1}^N \sum_{\beta=1}^N \euc{x_{\alpha}-x_{\beta}}\\
&=
-\euc{x_i}-\euc{x_j} + x_i \cdot x_j\\
&+ \frac{1}{N} \sum_{\beta=1}^N \euc{x_i} + \frac{1}{N} \sum_{\beta=1}^N \euc{x_{\beta}} - \frac{1}{N} \sum_{\beta=1}^N x_i \cdot x_{\beta}\\
&+ \frac{1}{N} \sum_{\alpha=1}^N \euc{x_{\alpha}} + \frac{1}{N} \sum_{\alpha=1}^N \euc{x_j} - \frac{1}{N} \sum_{\alpha=1}^N x_{\alpha} \cdot x_j\\
&- \frac{1}{N^2} \sum_{\alpha=1}^N \sum_{\beta=1}^N \euc{x_{\alpha}} - \frac{1}{N^2} \sum_{\alpha=1}^N \sum_{\beta=1}^N \euc{x_{\beta}}
+ \frac{1}{N^2} \sum_{\alpha=1}^N \sum_{\beta=1}^N x_{\alpha} \cdot x_{\beta}\\
\endaligned
$$
\normalsize
On the right-hand side above, averages in which the summation index does not appear in the summed terms simplify to
a single one of those terms, as in $\frac{1}{N} \sum_{k=1}^N a_m = a_m$. This leads to cancelation by pairs of all
squared terms. The remaining bilinear terms sum to the double-centered linear kernel:
\footnotesize
$$
\aligned
{\doublecentered{K}_E}_{i,j}
&=
x_i \cdot x_j
- \frac{1}{N} \sum_{\beta=1}^N x_i \cdot x_{\beta}
- \frac{1}{N} \sum_{\alpha=1}^N x_{\alpha} \cdot x_j
+ \frac{1}{N^2} \sum_{\alpha=1}^N \sum_{\beta=1}^N x_{\alpha} \cdot x_{\beta}\\
&=
{\doublecentered{K}_{\text{lin}}}_{i,j}
\endaligned
$$
\normalsize 
\boxenddemo
\begin{corollary}
The Euclidean and linear RKHS feature embeddings of a given data set are isometric. 
\label{corollary:EuclidRKHSIsometricToLinearRKHS}
\end{corollary}
\demo{Proof:}
This follows from Eq.~\ref{eq:metricFromGramMatrix} and Lemmas~\ref{lemma:doubleCenteringPreservesFeatureDistances},~\ref{lemma:doubleCenteredEuclideanGramsAreLinear}.
\boxenddemo

\subsection{Asymptotic linearity of the Gaussian kernel RKHS}
\label{section:AsymptoticLinearityGaussRKHS}

We will now state and prove the main result regarding the asymptotic metric structure of the RKHS feature spaces. 
We will rely on the notions of $\epsilon$-near-isometry and metric dissonance given in Definition~\ref{def:nearIsometry}.

\begin{definition}
a) Let $\epsilon > 0$. A mapping $T:X \rightarrow Y$\linebreak between metric spaces $X, Y$ is an {\em $\epsilon$-near-isometry} if 
$\vert d_Y(T(x_1), T(x_2)) - d_X(x_1, x_2) \vert \le \epsilon$ for all $x_1, x_2 \in X$,
and a {\em relative} $\epsilon$-near isometry if $\vert d_Y(T(x_1), T(x_2)) - d_X(x_1, x_2) \vert \le \epsilon\, \text{diam}(X)$.\\
b) The {\em metric dissonance} of $T$ is the smallest value of $\epsilon$ for which $T$ is an $\epsilon$-near-isometry;
this value is the least upper bound $\sup_{x_1, x_2 \in X} |d_Y(T(x_1), T(x_2)) - d_X(x_1,x_2)|$;\linebreak
the {\em relative metric dissonance} is the ratio of the dissonance to the diameter $\text{diam}_X = \sup_{x_1, x_2 \in X} d_X(x_1, x_2)$.
\label{def:nearIsometry}
\end{definition}
Our main result on the large-bandwidth asymptotics of Gaussian RBF RKHS geometry appears in Theorem~\ref{thm:GaussianPCAResidualDetails}. 
\begin{theorem}
For any finite data set, $X$, the Gaussian kernel RKHS feature representation of $X$ scaled (dilated) by $\sigma_X = d_X\sigma$ is relatively $\left ( \frac{\rho}{\sigma} \right )^2\!\! \eta$-near-isometric (Def.~\ref{def:nearIsometry}) 
to the linear kernel feature representation of $X$ as $\sigma \rightarrow \infty$, 
where $\sigma$ is the Gaussian bandwidth, $\rho$ is the representation eccentricity of Eq.~\ref{eq:rho_single_Gaussian_kernel},
and $\eta$ is the ratio of the diameter of $X$ to the minimum nonzero pairwise (Euclidean or
linear kernel-based) distance in $X$. 
In particular, the Gaussian and linear kernel RKHS representations of $X$ are asymptotically geometrically similar 
as $\sigma \rightarrow \infty$.
\label{thm:GaussianPCAResidualDetails}
\end{theorem}

\paragraph*{Notes}
\begin{enumerate}
\item Theorem~\ref{thm:GaussianPCAResidualDetails} is equivalent to the statement that the relative metric dissonance between the Gaussian and linear RKHS representations is at most $\left ( \frac{\rho}{\sigma} \right )^2\! \eta$.
\item Since the linear kernel Gram matrix consists of the Euclidean inner products, the corresponding canonical feature map is the
identity. Therefore, Theorem~\ref{thm:GaussianPCAResidualDetails} states that the Gaussian RKHS representation is asymptotic 
for large $\sigma$ to the original (linear) data representation scaled by $1/\sigma_X$.
\item The occurrence of $\sigma_X = d_X \sigma$ in Theorem~\ref{thm:GaussianPCAResidualDetails} derives from the scaling convention adopted in Eq.~\ref{eq:Gaussian_Gram_matrix}, that is, from the assumption that $\sigma$ is in standardized units. The scale factor would be simply $\sigma$ otherwise.
\end{enumerate}

\bigskip
\demo{Proof of Theorem~\ref{thm:GaussianPCAResidualDetails}:}
By Lemmas~\ref{lemma:doubleCenteringPreservesFeatureDistances},~\ref{lemma:doubleCenteredEuclideanGramsAreLinear}, 
we can assume without loss of generality that the Gram matrices 
$K = K_{G(\sigma)}$ and $L = K_{\text{lin}} = K_E$ have been double-centered. 

We will obtain the desired $O(\rho/\sigma)^2$ bound on the metric dissonance below,
by examining the details of the asymptotic dependence of the  
Gram matrix $\doublecentered{K}_{G(\sigma)}$ on $\sigma$, in order to identify a scale factor explicitly. 

We apply Eq.~\ref{eq:microDetailsDoubleCenteredK} to the Gaussian kernel of Eq.~\ref{eq:Gaussian_Gram_matrix}.
Linearizing the exponential $e^{-u}$ in the latter, we obtain 
Eq.~\ref{eq:microDetailsDoubleCenteredKsigma}, where $\alpha_{p,q} = |x_p - x_q|$
and $\sigma_X = d_X \sigma$; we keep first-order terms in the expansion only (quadratic in $\alpha_{p,q}$,\ $\sigma$) due to space limits.

\footnotesize
\begin{equation}
\aligned
{\doublecentered{K}_{G(\sigma)}}_{i,j} 
\approx 
- \frac{\alpha^2_{i,j}}{2\sigma^2_X}  
+ \frac{1}{N}\sum_{k=1}^N \frac{\alpha_{k,j}^2}{2\sigma^2_X} 
+ \frac{1}{N}\sum_{k=1}^N \frac{\alpha_{i,k}^2}{2\sigma^2_X} 
- \frac{1}{N^2}\sum_{k=1}^N \sum_{m=1}^N \frac{\alpha_{k,m}^2}{2\sigma^2_X} 
\label{eq:microDetailsDoubleCenteredKsigma}
\endaligned
\end{equation}
\normalsize
The right-hand side of Eq.~\ref{eq:microDetailsDoubleCenteredKsigma} is 
precisely $\frac{1}{\sigma^2_X} {\doublecentered{K}_E}_{i,j}$, where $\doublecentered{K}_E$ is the double-centered Euclidean Gram matrix.

The second-order expansion $e^{-u} \approx 1 - u + \frac{1}{2} u^2$ adds one term for each of the terms on the right-hand side of
Eq.~\ref{eq:microDetailsDoubleCenteredKsigma}. The second-order terms (quartic in $\alpha_{p,g}$,\ $\sigma$) are as in
Eq.~\ref{eq:microDetails2DoubleCenteredKsigma}.
\footnotesize
\begin{equation}
\aligned
&{\doublecentered{K}_{G(\sigma)}}_{i,j} - \frac{1}{\sigma^2_X}{\doublecentered{K}_E}_{i,j}\\
&\approx 
\frac{1}{2} \left (
\frac{\alpha^4_{i,j}}{4\sigma^4_X}  
- \frac{1}{N}\sum_{k=1}^N \frac{\alpha_{k,j}^4}{4\sigma^4_X} 
- \frac{1}{N}\sum_{k=1}^N \frac{\alpha_{i,k}^4}{4\sigma^4_X} 
+ \frac{1}{N^2}\sum_{k=1}^N \sum_{m=1}^N \frac{\alpha_{k,m}^4}{4\sigma^4_X} 
\right )
\label{eq:microDetails2DoubleCenteredKsigma}
\endaligned
\end{equation}
\normalsize
Using Eq.~\ref{eq:metricFromGramMatrix}, 
we find the squared distance difference between the Gaussian RKHS and the $\frac{1}{\sigma_X}$-scaled Euclidean RKHS:
\footnotesize
\begin{equation}
\aligned
&| x^{(i)} - x^{(j)} |^2_{\doublecentered{K}_{G(\sigma)}} \!\!-\ \frac{1}{\sigma^2_X}| x^{(i)} - x^{(j)} |^2_{E}\\
&=
{\doublecentered{K}_{G}}_{i,i} - \frac{1}{\sigma^2_X}{\doublecentered{K}_E}_{i,i}
+ {\doublecentered{K}_{G}}_{j,j} - \frac{1}{\sigma^2_X}{\doublecentered{K}_E}_{j,j}
-2 ({\doublecentered{K}_{G}}_{i,j} - \frac{1}{\sigma^2_X}{\doublecentered{K}_E}_{i,j})
\endaligned
\label{eq:microDetails3DoubleCenteredKsigma}
\end{equation}
\normalsize
Substituting Eq.~\ref{eq:microDetails2DoubleCenteredKsigma} 
in each of the three like summands in Eq.~\ref{eq:microDetails3DoubleCenteredKsigma} and simplifying
as in the proof of Lemma~\ref{lemma:doubleCenteringPreservesFeatureDistances},
we now obtain the second-order estimate in Eq.~\ref{eq:microDetails4DoubleCenteredKsigma}.
\footnotesize
\begin{equation}
\aligned
&| x^{(i)} - x^{(j)} |^2_{\doublecentered{K}_{G(\sigma)}} \!\!-\ \frac{1}{\sigma^2_X} | x^{(i)} - x^{(j)} |^2_{E}
\ \approx \ 
\frac{1}{8}
\left (
\frac{\alpha^4_{i,i}}{\sigma^4_X}
+\frac{\alpha^4_{j,j}}{\sigma^4_X}
-2\frac{\alpha^4_{i,j}}{\sigma^4_X}
\right )
\endaligned
\label{eq:microDetails4DoubleCenteredKsigma}
\end{equation}
\normalsize

Each of the $\alpha_{p,q}$ is bounded above by the diameter of the original data representation, $X$.
Since $\sigma_X = d_X \sigma$, each of the four parenthesized terms in Eq.~\ref{eq:microDetails4DoubleCenteredKsigma}
is bounded by $\frac{\rho^4}{\sigma^4}$, where $\rho$ is as in Eq.~\ref{eq:rho_single_Gaussian_kernel};
the entire right-hand side is bounded by $\frac{1}{2} \frac{\rho^4}{\sigma^4}$.
Equivalently, scaling both sides by $\sigma^2_X = d^2_X \sigma^2$ we have
\begin{equation}
\left \vert \sigma^2_X | x^{(i)} - x^{(j)} |^2_{\doublecentered{K}_{G(\sigma)}} \!\!-\ | x^{(i)} - x^{(j)} |^2_{E} \right \vert
\ \lesssim \  
\frac{1}{2} \frac{\rho^2}{\sigma^2} \text{diam}^2_X
\label{eq:scaledGramDifference}
\end{equation}
We extract the difference of the unsquared distances from the difference of their squares in Eq.~\ref{eq:scaledGramDifference} by
dividing by the distance sum $\sigma^2_X | x^{(i)} - x^{(j)} |_{\doublecentered{K}_{G(\sigma)}} \!\! +\ | x^{(i)} - x^{(j)} |_{E}$. Ignoring the Gaussian distance summand altogether, this yields the loose bound in Eq.~\ref{eq:scaledDistanceDifference}, where $\eta = \text{diam}(X)\ / \min\limits_{x_1 \ne x_2}{d_X(x_1, x_2)}$.
\begin{equation}
\left \vert \sigma^2_X | x^{(i)} - x^{(j)} |_{\doublecentered{K}_{G(\sigma)}} \!\!-\ | x^{(i)} - x^{(j)} |_{E} \right \vert
\ \lesssim \  
\frac{1}{2} \frac{\rho^2}{\sigma^2} \eta\, \text{diam}_X
\label{eq:scaledDistanceDifference}
\end{equation}
Doubling the multiplicative factor in Eq.~\ref{eq:scaledGramDifference} to account for higher-order terms,
it follows that the RKHS embeddings associated with
${\sigma^2_X} \doublecentered{K}_{G(\sigma)}$ and $\doublecentered{K}_{\text{lin}}$ are 
asymptotically relatively $\frac{\rho^2}{\sigma^2}\eta$-near-isometric (Def.~\ref{def:nearIsometry}).
Theorem~\ref{thm:GaussianPCAResidualDetails} follows. 
\boxenddemo

\subsubsection*{Preservation of angles}

Corollary~\ref{corollary:anglePreservation} shows that asymptotic similarity of the Gaussian kernel and linear kernel RKHS in a metric sense as in Theorem~\ref{thm:GaussianPCAResidualDetails} extends to asymptotic preservation of angles. 

\begin{corollary}
Consider any three data points, with linear RKHS representations $x_i$ and Gaussian RKHS representations $y_i$, $i=1,2,3$. 
Let $\theta_{i,j,k}$ and $\phi_{i,j,k}$ denote the angles defined by the $x_i$ and $y_i$ triples, respectively, with $x_2$
and $y_2$ as the respective vertices. Then $\theta_{i,j,k} - \phi_{i,j,k} = O(\rho/\sigma)$ as $\sigma \rightarrow \infty$. 
\label{corollary:anglePreservation}
\end{corollary}
\demo{Proof}
In light of Theorem~\ref{thm:GaussianPCAResidualDetails}, the bound 
$\vert \frac{a + \epsilon}{b - \epsilon} - \frac{a}{b} \vert \le 3 \frac{\epsilon}{b}$
for any real numbers $a, b$ with $|a| \le b$ and any $\epsilon < b/3$
shows that the cosine difference 
$\frac
{
(\tilde{y}_2-\tilde{y}_1) \cdot_Y (\tilde{y}_3-\tilde{y}_1)
}
{
|\tilde{y}_2-\tilde{y}_1| |\tilde{y}_3-\tilde{y}_1|
} 
- 
\frac
{
(\tilde{x}_2-\tilde{x}_1) \cdot_X (\tilde{x}_3-\tilde{x}_1)
}
{
|\tilde{x}_2-\tilde{x}_1| |\tilde{x}_3-\tilde{x}_1|
}$ 
is also $O \left ( \frac{\rho}{\sigma} \right )^2$.
This implies that the angles formed by the given triple of data points will differ between the two RKHS by $O \left ( \frac{\rho}{\sigma} \right )$; the angles will be equal in the limit as $\sigma \rightarrow \infty$.
\boxenddemo

\subsection{Asymptotic linearity of Gaussian kernel PCA}

In light of the results of section~\ref{section:AsymptoticLinearityGaussRKHS},  
one expects that Gaussian kernel PCA, which hinges directly on the geometry of the RKHS
feature embeddings, will also behave near-linearly for large bandwidths. 
Theorem~\ref{thm:GaussPCASpectralLimit} describes this behavior in the
generic case (probability $1$) of simple linear eigenvalues. 

\begin{theorem}
Assume the linear Gram matrix $\doublecentered{K}_{\text{lin}}$ has no repeated eigenvalues.
The scaled Gaussian kernel PCA eigenvalues $\sigma^2_X \lambda_i$, eigenprojections 
of the Gaussian Gram matrix, and scaled Gaussian kernel PCA principal components $\sigma_X \Phi(x_i) \cdot V^k$ 
converge to their linear PCA counterparts as $\sigma \rightarrow \infty$.
The eigenvalues of $\doublecentered{K}_{G(\sigma)}$ satisfy $\lambda_i = O\left ( \frac{\rho}{\sigma} \right )^2$,
where $\rho$ is the data eccentricity (Eq.~\ref{eq:rho_single_Gaussian_kernel}).
\label{thm:GaussPCASpectralLimit}
\end{theorem}
\demo{Proof of Theorem~\ref{thm:GaussPCASpectralLimit}:}
As described in section~\ref{section:KernelPCA}, the kernel PCA projection is determined by the eigenvectors of the 
double-centered Gaussian Gram matrix, $\doublecentered{K}_{G(\sigma)}$, together with the data feature vectors in the RKHS.
The expression $e^{\frac{|x_i-x_j|^2}{2 d^2_X \sigma^2}}$ that defines the entries of the uncentered matrix $K_{G(\sigma)}$ (Eq.~\ref{eq:Gaussian_Gram_matrix}) 
is analytic in $\epsilon = \frac{1}{d_X \sigma} = \frac{1}{\sigma_X}$ for all $i, j$. 
Analyticity in $\epsilon$ is retained by double-centering, as this step involves only multiplication by the
constant centering matrix, $H$.
These facts enable the use of the analytic perturbation theory of linear operators~\cite{KatoPerturbationTheoryBook1995} 
(chapter II, section 1). 

The centered Gaussian Gram matrix, $\doublecentered{K}_{G(\sigma)}$, approaches the identically zero matrix as $\epsilon \rightarrow 0$. 
In order to avoid a perturbation problem that is singular at $\epsilon = 0$, we use Eq.~\ref{eq:microDetailsDoubleCenteredKsigma} and Eq.~\ref{eq:microDetails2DoubleCenteredKsigma} to express $\doublecentered{K}_{G(\sigma)}$ 
as a scaling of the perturbed centered Euclidean Gram matrix, $\doublecentered{K}_E$, 
as in Eq.~\ref{eq:Gaussian_Gram_matrix_as_scaled_nonsingular_perturbation}. 
\begin{equation}
\doublecentered{K}_{G(\sigma)} = \epsilon^2 \left ( \doublecentered{K}_E + O(\epsilon^2) \right )
\label{eq:Gaussian_Gram_matrix_as_scaled_nonsingular_perturbation}
\end{equation}
By Lemma~\ref{lemma:doubleCenteredEuclideanGramsAreLinear}, we can replace $\doublecentered{K}_E$ in Eq.~\ref{eq:Gaussian_Gram_matrix_as_scaled_nonsingular_perturbation} by the double-centered linear Gram matrix, 
$\doublecentered{K}_{\text{lin}}$.
We can now consider the perturbation problem for the rescaled symmetric matrix $\frac{1}{\epsilon^2} \doublecentered{K}_{G(\sigma)} = \doublecentered{K}_{\text{lin}} + O(\epsilon^2)$,
which is non-singular at $\epsilon = 0$.

Since $\doublecentered{K}_{\text{lin}}$ has no repeated eigenvalues, the eigenvalues of 
the rescaled perturbation $\doublecentered{K}_{\text{lin}} + O(\epsilon^2)$ are simple 
for all sufficiently small $\epsilon$ (exceptional points are isolated~\cite{KatoPerturbationTheoryBook1995}, II.1.1).
The perturbed matrix is symmetric by Eq.~\ref{eq:Gaussian_Gram_matrix_as_scaled_nonsingular_perturbation} (therefore normal), so its individual eigenvalues, eigenprojections, and individual eigenvectors are analytic in $\epsilon$ and converge to the respective eigenvalues and eigenvectors of the linear kernel $\doublecentered{K}_{\text{lin}}$ as $\epsilon \rightarrow 0$~\cite{KatoPerturbationTheoryBook1995} (chapter II, section 1). In the case of repeated eigenvalues of $\doublecentered{K}_{\text{lin}}$ (which we do not consider), the sum of the eigenprojections of the perturbed matrix associated with a given repeated eigenvalue still converges to the corresponding eigenprojection of the latter, but individual eigenvectors may not~(\!\cite{KatoPerturbationTheoryBook1995}, II.1.6, incl.\ Remark 1.11).

By Eq.~\ref{eq:Gaussian_Gram_matrix_as_scaled_nonsingular_perturbation}, the eigenvalues of the Gaussian Gram matrix, 
$\doublecentered{K}_{G(\sigma)}$, 
are scalings $\lambda_i = \epsilon^2 \mu_i = \frac{1}{\sigma^2_X} \mu_i$ of the eigenvalues of 
$\doublecentered{K}_{\text{lin}} + O(\epsilon^2)$ by $1/\sigma^2_X$. 
The eigenvectors of $\doublecentered{K}_{G(\sigma)}$
align with the very same $\alpha^i$ as for $\doublecentered{K}_{\text{lin}} + O(\epsilon^2)$;
for the purpose of kernel PCA, however, the eigenvectors are normalized as in section~\ref{section:KernelPCA} so that
$\lambda_i \alpha^i \cdot \alpha^i = 1$, while the linear PCA eigenvector normalization condition is 
$\mu_i \alpha^i \cdot \alpha^i = \sigma^2_X \lambda_i \alpha^i_{\text{lin}} \cdot \alpha^i_{\text{lin}} = 1$. Therefore, 
as $\epsilon \rightarrow 0$ (as $\sigma \rightarrow \infty$), the scaled Gaussian eigenvalues $\sigma^2_X \lambda_i$ 
will converge to the eigenvalues of $\doublecentered{K}_{\text{lin}}$, and the scaled Gaussian eigenvectors $\frac{1}{\sigma_X} \alpha^i$
will converge to the respective linear eigenvectors $\alpha^i_{\text{lin}}$ (i.e., the total eigenprojections will converge, as discussed above). 

Taking into account that $\doublecentered{K}_{G(\sigma)}$ is asymptotic to $\frac{1}{\sigma^2_X} \doublecentered{K}_{\text{lin}}$ 
for large $\sigma$ as implied by Eq.~\ref{eq:Gaussian_Gram_matrix_as_scaled_nonsingular_perturbation} (and Lemma~\ref{lemma:doubleCenteredEuclideanGramsAreLinear}),
it follows from the above that the scaled Gaussian kernel PCA principal components 
$\sigma_X \Phi(x_i) \cdot V^k = \sigma_X \sum_{j=1}^N \alpha^k_j K(x_j, x_i)$ 
(section~\ref{section:KernelPCA}) will converge to the corresponding linear kernel PCA principal components. 

The Courant-Fischer
minimax principle bounds the eigenvalues of a positive semi-definite symmetric $K$ by the ratios $\frac{x^T K x}{x^T x}$. 
Arguing as in the proof of Lemma~\ref{thm:GaussianPCAResidualDetails},
the exponential arguments $\frac{|x_i-x_j|^2}{2\sigma^2_X}$ in $K_{G(\sigma)}$ are uniformly $O\left (  \frac{\rho}{\sigma} \right )^2$.
It follows that the eigenvalues of $\doublecentered{K}_{G(\sigma)}$ are of size $O\left (  \frac{\rho}{\sigma} \right )^2$ as $\sigma \rightarrow \infty$. This completes the proof of Theorem~\ref{thm:GaussPCASpectralLimit}.
\boxenddemo

\section{Experimental Evaluation}
\label{section:exp}

\subsection{Experimental methodology}
\label{section:expMethodology}

\subsubsection{Data sets}
We evaluate the statements of Theorems~\ref{thm:GaussianPCAResidualDetails} and~\ref{thm:GaussPCASpectralLimit} for
a sample collection of $30$ data sets from OpenML~\cite{OpenML2013} (CC BY 4.0 license, \url{https://creativecommons.org/licenses/by/4.0/}). Target labels are ignored; one-hot numerical encoding is used for
categorical attributes only. Data set names and characteristics appear in Table~\ref{table:dataSets}.
The selected data sets span a wide range of eccentricity values $\rho$ (Eq.~\ref{eq:rho_single_Gaussian_kernel}), from a minimum of $1.15$ for \verb|splice|, to a maximum of over $1800$ for \verb|pc3|.

\begin{table}
\caption{Data set characteristics. Parenthesized numbers are attribute counts after one-hot numerical encoding of categorical attributes only. Names of starred (*) data sets are abbreviated:
blood-transfusion-service-center, climate-model-simulation-crashes,
one-hundred-plants-margin, one-hundred-plants-shape, one-hundred-plants-texture.}

\begin{tabular}{lccc}
\toprule
Data set &Examples &Attributes &Eccentricity, $\rho$\\
&&&(Eq.~\ref{eq:rho_single_Gaussian_kernel})\\
\hline
splice &3190 &60 (287) &1.15\\ 
tic-tac-toe &958 &9 (27) &1.22\\ 
wdbc &569 &30 &10.52\\
diabetes &768 &8 &8.41\\
wine &178 &13 &5.0\\
balloon &2001 &1 &14.83\\
cpu &209 &7 (36) &8.92\\ 
boston &506 &13 (22) &3.68\\ 
stock &950 &9 &2.55\\
cloud &108 &5 (9) &4.67\\ 
credit-g &1000 	&20 (61) 	&10.34\\ 
blood-transfusion-sc* &748 	&4 	&16.29\\
monks-problems-2 &601 	&6 (17)	&1.22\\ 
steel-plates-fault &1941 	&33 	&10.4\\
kr-vs-kp &3196 	&36 	(73) &1.63\\ 
qsar-biodeg &1055 	&41 	&8.5\\
phoneme &5404 	&5 	&2.57\\
ozone-level-8hr &2534 	&72 	&4.79\\
hill-valley &1212 	&100 	&41.42\\
kc2 &522 	&21 	&502.54\\
climate-model-sc* &540 	&20 	&3.17\\
spambase &4601 	&57 	&106.69\\
kc1 &2109 	&21 	&252.8\\
ilpd 				&583 	&10 (11) &33.4\\ 
pc1 					&1109 	&21 &608.41\\
pc3 					&1563 	&37 	&1806.03\\
one-hund-pl-margin* 	&1600 	&64 	&2.98\\
one-hund-pl-shape*	&1600 	&64 &6.6\\
one-hund-pl-texture*	&1599 	&64 &3.37\\
banknote-authentication 		&1372 	&4 	&3.87\\
\bottomrule
\end{tabular}
\label{table:dataSets}
\end{table}

\subsubsection{Unnormalized evaluation measures (used as in~\ref{section:normalizationOfMetrics})}
\label{section:evalMetrics}

\noindent
{\em RKHS metric convergence:}
To validate Theorem~\ref{thm:GaussianPCAResidualDetails}, we track metric dissonance 
$\max_{x_1, x_2 \in X} |\sigma_X d_G(x_1,x_2) - d_L(x_1,x_2)|$ (Def.~\ref{def:nearIsometry}) 
of the natural mapping between the RKHS corresponding to the scaled Gaussian and linear Gram matrices
($\sigma^2_X K_{G(\sigma)}$ and $K_{\text{lin}}$). 
We compute dissonance as the uniform ($L^{\infty}$) norm of the difference between the 
pairwise distance matrices, which we obtain from the Gram matrices via Eq.~\ref{eq:distSquaredRKHS}. 

\noindent
{\em Eigenvalue convergence:}
To validate Theorem~\ref{thm:GaussPCASpectralLimit}, we track the eigenvalues and eigenvectors 
of the Gaussian Gram matrix, which define the Gaussian PCA projections as described in section~\ref{section:KernelPCA}. 
We gauge the discrepancy between the scaled Gaussian and linear eigenvalue arrays by the uniform norm (maximum) 
of their difference. 

\noindent
{\em Eigenprojection convergence:}
We estimate the degree of misalignment or ``dissonance'' between total eigenprojections using Algorithm~\ref{alg:eigenprojection-dissonance}, where $v[1...L]$ and $w[1...L]$ are the Gaussian and linear eigenvector arrays, $\lambda[1...L]$ are the $k$ largest linear eigenvalues in descending order,
for some $k$ (which is less than $L$ if one or more eigenvalues are repeated), 
and $\tau$ is a numerical threshold for equality. 
Varying $k$ allows us to consider the dependence of convergence
behavior on eigenvalue rank.
In Alg.~\ref{alg:eigenprojection-dissonance}, we use the \verb|SciPy|~\cite{2020SciPy-NMeth} function \verb|PR_ANGLES|,
which returns the principal angles~\cite{KnyazevArgentatiPrincipalAngles2002},~\cite{knyazev2012principal} 
between the subspaces spanned by two sets of vectors.

\begin{figure}[h]
	\begin{center}
		\includegraphics [width=0.51\columnwidth, clip=true, trim=1mm 4mm 0mm 10mm]
		{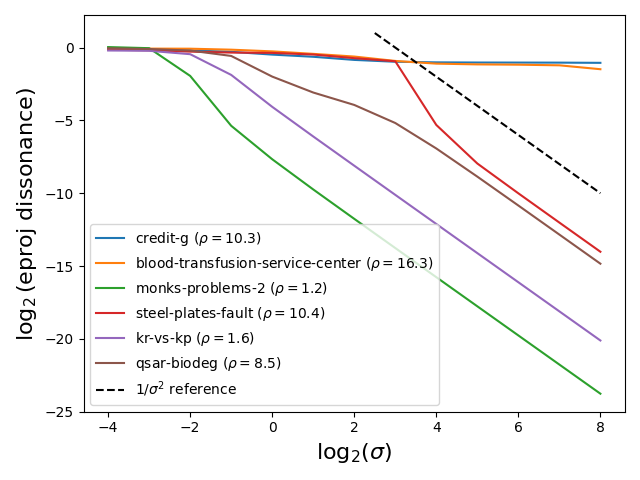}
		\includegraphics [width=0.48\columnwidth, clip=true, trim=10mm 4mm 0mm 10mm]
		{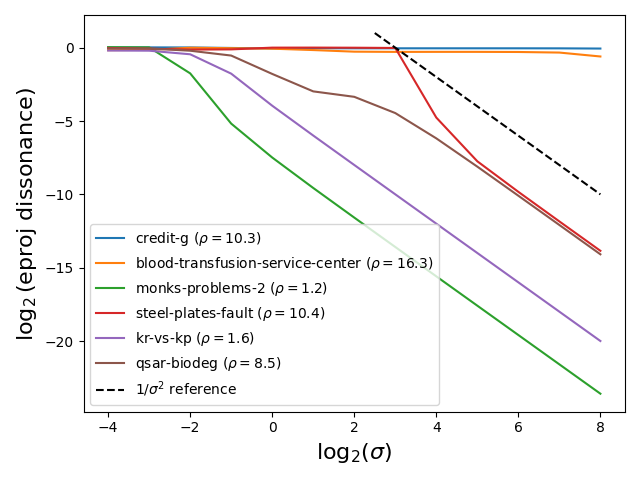}
	\end{center}
	\caption{Sample eigenprojection dissonance results for different aggregation operators: mean (left), max (right).
	The two aggregation operators (Alg.~\ref{alg:eigenprojection-dissonance}, line~\ref{line:aggregation}) typically yield qualitatively similar results, as shown here.}
	\label{fig:eproj_diss_aggr_comparison}
\end{figure}

Algorithm~\ref{alg:eigenprojection-dissonance} groups 
eigenvalues that differ by less than the threshold $\tau$ 
and returns an aggregate of the principal angles between the corresponding paired eigenspaces 
for each distinct linear eigenvalue. Smaller principal angles reflect 
closer agreement between eigenprojections.
We use $10^{-6}$ as the default value of the threshold $\tau$; values of $10^{-9}$ and $10^{-12}$ gave the same results. 
One might instead use the minimum difference between distinct
linear eigenvalues as the threshold, but limited-precision arithmetic may prevent reliably
determining true equality of eigenvalues. 
We use the mean as the default aggregation operator on line~\ref{line:aggregation} of Alg.~\ref{alg:eigenprojection-dissonance}; 
the maximum yields qualitatively similar results (e.g., Fig.~\ref{fig:eproj_diss_aggr_comparison}).

\begin{algorithm}
\caption{Eigenprojection dissonance. See section~\ref{section:evalMetrics}.}\label{alg:eigenprojection-dissonance}
\begin{algorithmic}[1]
\Function{EigenDist}{$v[1 \cdots L]; w[1 \cdots L]; \lambda[1 \cdots L]; \tau$}\\
\Comment{$v$: Gauss evecs; $w$: lin evecs; $\lambda$: lin evals; $\tau$: threshold}
    \State $\theta = [\ ]$
    \State $i_0 = 1$
    \While{$i_0 \le L$}
        \State $i = i_0$
        \While {$i \le L$ and $\left |\frac{\lambda[i]-\lambda[i_0]}{\lambda[i_0]} \right | < \tau$}
            \State $i = i + 1$
        \EndWhile
        \State $\theta$.append ( max $\vert$ {\Call{pr\_angles}{$v[i_0:i], w[i_0:i]$}} $\vert$ )
        \State $i_0 = i$
    \EndWhile
    \State \Return \Call{aggr}{$\theta$}	\label{line:aggregation}
\EndFunction
\end{algorithmic}
\end{algorithm}

\subsubsection{Normalized evaluation measures}
\label{section:normalizationOfMetrics}
The measures of section~\ref{section:evalMetrics} are normalized before use. This protocol aligns, in particular, with
the use of the relative version of the metric dissonance in Theorem~\ref{thm:GaussianPCAResidualDetails}.
We report the $\log_2$ value of each evaluation measure 
(RKHS metric dissonance, eigenvalue difference, eigenprojection dissonance), 
shifting the result to $0$ at $\sigma = 0$ (Eq.~\ref{eq:metricNormalization}, where dependence on the data has been
suppressed).
\begin{equation}
f_{\text{norm}}(\sigma) = \log_2 f(\sigma) - \log_2 f(0)
\label{eq:metricNormalization}
\end{equation} 
Eq.~\ref{eq:metricNormalization} corresponds to normalizing the measure, $f$, by dividing by $f(0)$. The use of a logarithmic scale enables capturing a wide range of orders of magnitude, and facilitates comparisons between data sets. 

The normalized (relativized) measures appear below. The symbol $I$ in the denominator of Eq.~\ref{eq:normalizedEigenprojectionDissonance}
denotes the identity matrix, the columns of which form an eigenvector basis for the matrix $\sigma^2_X K_{G(\sigma)}$ 
at $\sigma=0$
($\sigma_X = d_X \sigma$; c.f.,~Eqs.~\ref{eq:Gaussian_Gram_matrix},~\ref{eq:metricNormalization}).

\medskip
\paragraph*{RKHS metric dissonance}

\begin{equation}
\log_2 \frac{\max_{x_1, x_2 \in X} |\sigma_X d_{G(\sigma)}(x_1,x_2) - d_L(x_1,x_2)|}{\text{diam}_L(X)}
\label{eq:normalizedRKHSMetricDissonance}
\end{equation}

\paragraph*{Norm of eigenvalue difference}

\begin{equation}
\log_2 \frac{\Vert \sigma^2_X \lambda_{G(\sigma)} - \lambda_L \Vert_{\infty}}{\Vert \lambda_L \Vert_{\infty}}
\label{eq:normalizedEigenvalueDifference}
\end{equation}

\paragraph*{Eigenprojection dissonance (see Alg.~\ref{alg:eigenprojection-dissonance})}

\begin{equation}
\log_2 \frac{\text{EigenDist}(v,w,\lambda_L)}{\text{EigenDist}(I,w,\lambda_L)}
\label{eq:normalizedEigenprojectionDissonance}
\end{equation}

\subsubsection{Convergence onset}
\label{section:convergenceOnset}
We compute a convergence-onset bandwidth for each measure as in~\cite{AlvarezGaussianCKA2022}, corresponding to a boundary
between transient and asymptotic bandwidth regimes; see below. 

We begin by establishing an approximate $1/\sigma^2$ asymptote for the measure in question, by extrapolating from the measure's observed value at the largest reliable tested bandwidth, $\sigma = 2^7$; experiments at bandwidths higher than $2^7$ can incur numerical error. The convergence onset bandwidth, $\sigma^*_0 = 2^{p^*_0}$, is defined as the smallest tested bandwidth such that the given measure deviates by less than $25\%$ from its approximate $1/\sigma^2$ asymptote, for all bandwidths $\ge \sigma^*_0$. If no
such bandwidth exists, we use $2^8 = 256$ as a lower bound on the onset bandwidth. We report $\log_2 \sigma^*_0 = p^*_0$ instead of $\sigma^*_0$ in some cases, in order to facilitate visual comparisons.

\subsubsection{Experimental procedure}
\label{section:experimentalProcedure}
For each data set, we compare the results of linear kernel PCA and Gaussian kernel PCA, for each one of the standardized Gaussian bandwidths $\sigma = 2^p$, for $p=-4, -3, -2, \cdots, 8$. Standardized bandwidths are scaled by the median pairwise distance between data examples, as in Eq.~\ref{eq:Gaussian_Gram_matrix}. That is to say, we consider bandwidths between $1/16$ and $256$ times the
median distance between examples (Eq.~\ref{eq:Gaussian_Gram_matrix}). For bandwidths of $256$ or higher, the small magnitudes of the eigenvalues of lesser rank lead to a loss of numerical precision in the differences $\sigma^2_X \lambda_{G(\sigma)} - \lambda_L$, making it difficult to obtain reliable numerical results. Results for bandwidths larger than $128$ are therefore not considered in the analysis of experimental results below, though the value $\sigma = 256$ is included in plots.

\subsubsection{Hardware and software platforms}
Experiments were performed on a workstation with an Intel i9-7920X (12 core, 2.9 GHz) processor and 128GB RAM, under Ubuntu 20.04.4 LTS (GNU public license). No GPU acceleration was used. 
Total compute time for one run of the experiments presented in the paper was approximately $100$ minutes.
Several preliminary partial runs were made for hyperparameter selection.
CO$_2$ emitted per full run as reported by CodeCarbon~\cite{CodeCarbonDOI} was $\approx 17$g; energy
was $46$Wh.

We used the kernel PCA implementation provided in \verb|scikit-learn|~\cite{scikit-learn} (BSD 3 license), with additional functions in NumPy~\cite{NumPy2020} (BSD 3 license), 
SciPy~\cite{2020SciPy-NMeth} (BSD 3 license), 
Matplotlib~\cite{Matplotlib2007} (PSF license, \url{https://docs.python.org/3/license.html}), and 
\verb|pandas|~\cite{reback2020pandas, mckinney-proc-scipy-2010} (BSD 3 license).

\subsection{Experimental results}
\label{section:expResults}

\subsubsection{RKHS metric convergence}

\begin{figure}[h]
	\begin{center}
		\includegraphics [width=0.8\columnwidth, clip=true, trim=0mm 4mm 0mm 2mm]
		{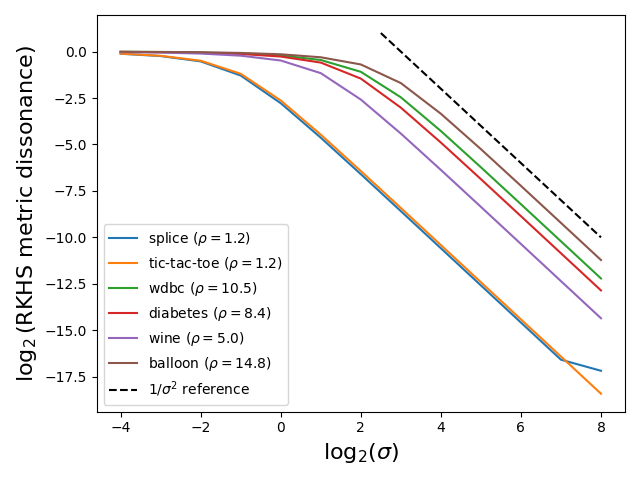}
	\end{center}
	\caption{RKHS metric convergence, six sample data sets;
	results are similar for other data sets. 
	Horizontal axis shows $\log_2$ standardized Gaussian bandwidth. 
	Vertical axis shows normalized $\log_2$ RKHS metric dissonance (Eq.~\ref{eq:normalizedRKHSMetricDissonance}).
	$1/\sigma^2$ convergence as $\sigma \rightarrow \infty$ is observed, as evidenced by an
	asymptotic slope of $-2$ for large $\sigma$. 
	Dissonance increases with data eccentricity, $\rho$; curves for two lowest eccentricity data sets of
	are nearly indistinguishable; wdbc value at $\log_2 \sigma = 8$ (lower right) reflects numerical error;
	see sections~\ref{section:convergenceOnset},~\ref{section:experimentalProcedure}.}
	\label{fig:RKHS_diss}
\end{figure}

Fig.~\ref{fig:RKHS_diss} shows that metric dissonance (Eq.~\ref{eq:normalizedRKHSMetricDissonance}) 
between the RKHS corresponding to the scaled Gaussian PCA Gram matrix 
$\sigma^2_X K_{G(\sigma)}$ and the linear PCA Gram matrix $K_{\text{lin}}$ decreases like $1/\sigma^2$ as $\sigma \rightarrow \infty$
(asymptotic slope of $-2$ in logarithmic coordinates), 
precisely in accordance with Theorem~\ref{thm:GaussianPCAResidualDetails}.

\begin{figure}[h]
	\begin{center}
		\includegraphics [width=0.8\columnwidth, clip=true, trim=0mm 4mm 0mm 2mm]
		{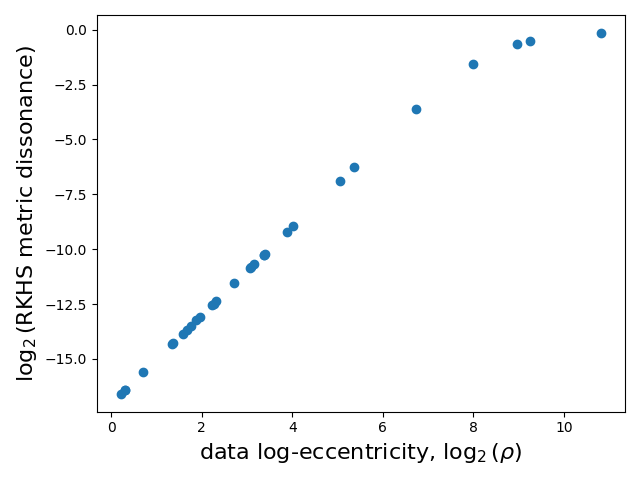}
	\end{center}
	\caption{Large-bandwidth RKHS metric dissonance (Eq.~\ref{eq:normalizedRKHSMetricDissonance}) increases approximately quadratically with data eccentricity, $\rho$ (Eq.~\ref{eq:rho_single_Gaussian_kernel});
	plot uses $\log_2$ values on both axes.
	Results shown for $\sigma=2^7$, and $30$ data sets. Slope of least-squares linear fit in log coordinates is $1.74$;
	coefficient of determination $r^2 = 0.98$. Norm of eigenvalue difference (not shown) also grows 
	near-quadratically with $\rho$ (slope $1.71$, $r^2 = 0.97$).}
	\label{fig:rho_vs_dissons}
\end{figure}

Also, for each fixed bandwidth $\sigma \gg 1$, metric dissonance increases
with data eccentricity, $\rho$ (Eq.~\ref{eq:rho_single_Gaussian_kernel}). 
Fig.~\ref{fig:rho_vs_dissons} confirms the latter point, showing, further, that metric dissonance
increases approximately quadratically with $\rho$. 
This behavior is consistent with the $C \left (\frac{\rho}{\sigma} \right )^2$
bound in Theorem~\ref{thm:GaussianPCAResidualDetails}. Table~\ref{table:convergenceOnsetsVsRhos}
(leftmost of the three $\sigma^*_0$ columns) shows that the onset of $1/\sigma^2$ convergence occurs
at $\sigma \approx \rho$, again in line with the $O \left (\frac{\sigma}{\rho} \right )^2$
bound of Theorem~\ref{thm:GaussianPCAResidualDetails}.

\subsubsection{Eigenvalue convergence}
\label{section:eval_convergence}

\begin{figure}[h]
	\begin{center}
		\includegraphics [width=0.8\columnwidth, clip=true, trim=0mm 3mm 0mm 2mm]
		{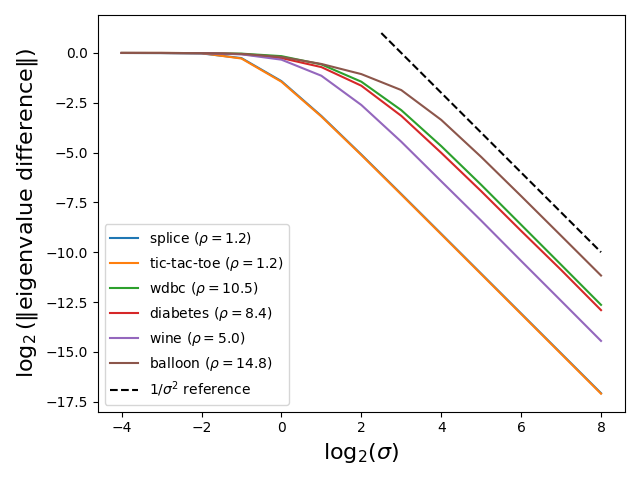}
	\end{center}
	\caption{Eigenvalue convergence, six sample data sets. Horizontal axis shows $\log_2$ 
	standardized Gaussian bandwidth. 
	Vertical axis shows normalized eigenvalue difference norm
	(Eq.~\ref{eq:normalizedEigenvalueDifference}).
	Asymptotic $-2$ slope demonstrates $1/\sigma^2$ convergence
	as $\sigma\rightarrow \infty$.
	Difference norm increases with data eccentricity, $\rho$ (Eq.~\ref{eq:rho_single_Gaussian_kernel}); 
	curves for the two data sets of
	lowest eccentricity are indistinguishable.}
	\label{fig:eval_diff}
\end{figure}

Fig.~\ref{fig:eval_diff} shows that the relative norm $\Vert \sigma^2_X \lambda_{G(\sigma)} - \lambda_L \Vert/\Vert \lambda_L \Vert$ 
of the difference
between the scaled Gaussian PCA eigenvalue array $\sigma^2_X \lambda_{G(\sigma)}$ and the linear PCA eigenvalue array $\lambda_L$
decreases like $1/\sigma^2$ as $\sigma \rightarrow \infty$, precisely in accordance with Theorem~\ref{thm:GaussPCASpectralLimit}.
For a fixed bandwidth $\sigma \gg 1$, eigenvalue difference increases approximately quadratically with data eccentricity, $\rho$: 
for $\sigma = 2^7$, for example, the slope of least-squares linear fit to the $\rho$-vs-difference graph in log coordinates is $1.74$, and the coefficient of determination $r^2$, is $0.98$. Table~\ref{table:convergenceOnsetsVsRhos} (middle $\sigma^*_0$ column) shows that onset of $1/\sigma^2$ convergence (section~\ref{section:convergenceOnset}) occurs at $\sigma \approx \rho$ across data sets.

\subsubsection{Eigenprojection convergence}

\begin{figure}[h]
	\begin{center}
		\includegraphics [width=0.8\columnwidth, clip=true, trim=0mm 4mm 0mm 3mm]
		{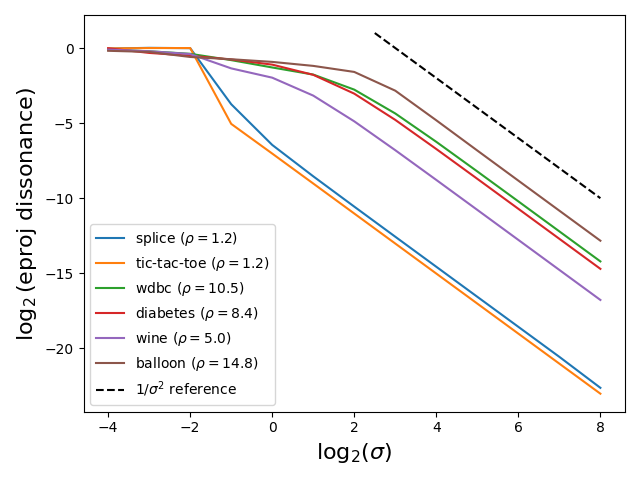}
	\end{center}
	\vspace{-0.2cm}
	\caption{Eigenprojection convergence, top eigenvalue only, six sample data sets. 
	Horizontal axis shows standardized Gaussian bandwidth. 
	Vertical axis shows normalized eigenprojection dissonance (Eq.~\ref{eq:normalizedEigenprojectionDissonance}).
	Asymptotic $-2$ slope in logarithmic coordinates confirms $1/\sigma^2$ convergence. 
	Dissonance increases with eccentricity, $\rho$ (Eq.~\ref{eq:rho_single_Gaussian_kernel}); 
	see Fig.~\ref{fig:rho_vs_eprojdissons}.}
	\label{fig:eproj_diss_1}
\end{figure}

\begin{figure}[h]
	\begin{center}
		\includegraphics [width=0.8\columnwidth, clip=true, trim=0mm 3mm 0mm 2mm]
		{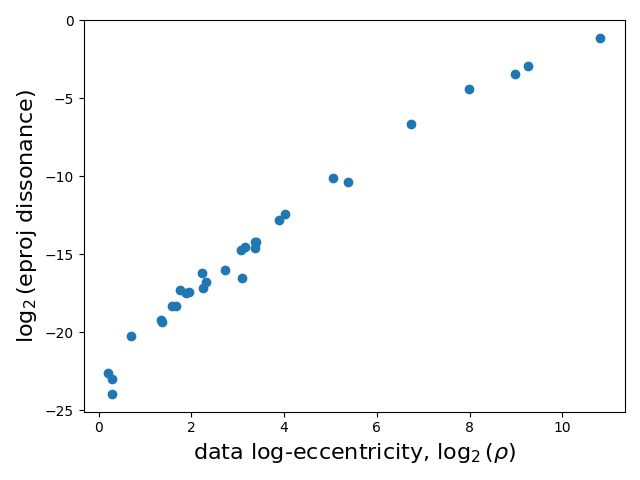}
	\end{center}
	\vspace{-0.3cm}
	\caption{Large-$\sigma$ normalized eigenprojection dissonance (Eq.~\ref{eq:normalizedEigenprojectionDissonance}) 
	vs.\ $\log_2$ data eccentricity $\rho$ (Eq.~\ref{eq:rho_single_Gaussian_kernel}), top eigenvalue only. 
	Results shown for $\sigma=2^7$, and $30$ data sets. 
	Dissonance increases with data eccentricity, $\rho$.
	Slope of least-squares linear fit is $2.05$;
	coefficient of determination, $r^2$, is $0.97$.}
	\label{fig:rho_vs_eprojdissons}
\end{figure}

Convergence of the Gaussian eigenprojections toward their linear counterparts is found to be stratified by eigenvalue rank.
For the top (largest) eigenvalue, the observed $1/\sigma^2$ convergence (Fig.~\ref{fig:eproj_diss_1}) mirrors that of RKHS metric dissonance (Fig.~\ref{fig:RKHS_diss}) and eigenvalue difference (Fig.~\ref{fig:eval_diff}). In this case,  eigenprojection dissonance at the largest tested bandwidth $\sigma = 2^7$
correlates strongly with data eccentricity, $\rho\ (r^2 = 0.97)$; see Fig.~\ref{fig:rho_vs_eprojdissons}. 
Convergence onset also mirrors
$\rho$; see Table~\ref{table:convergenceOnsetsVsRhos} 
(right column). These results are very much in line with Theorem~\ref{thm:GaussPCASpectralLimit}.

\begin{figure}[h]
	\begin{center}
		\includegraphics [width=0.51\columnwidth, clip=true, trim=3mm 4mm 0mm 2mm]
		{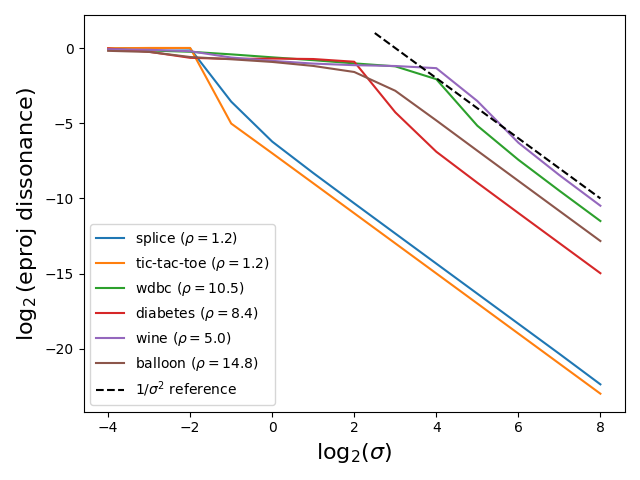}
		\includegraphics [width=0.48\columnwidth, clip=true, trim=10mm 4mm 2mm 2mm]
		{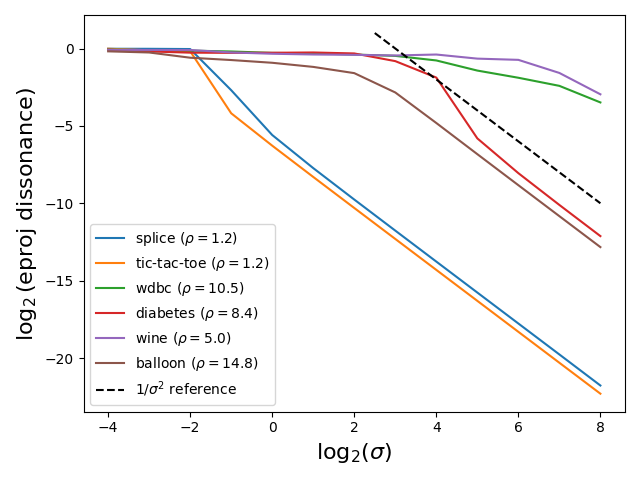}
	\end{center}
	\vspace{-0.3cm}
	\caption{Eigenprojection convergence: top two eigenvalues (left), top five eigenvalues (right). 
	Vertical axis shows normalized eigenprojection dissonance (Eq.~\ref{eq:normalizedEigenprojectionDissonance}).
	Convergence onset bandwidth (section~\ref{section:convergenceOnset}) increases as more 
	eigenvalues are considered. 
	Dissonance is no longer monotone in eccentricity $\rho$ (Eq.~\ref{eq:rho_single_Gaussian_kernel}), 
	unlike the single eigenvalue case (c.f., Fig.~\ref{fig:eproj_diss_1}).}
	\label{fig:eproj_diss_n}
\end{figure}

For multiple eigenvalues as in Fig.~\ref{fig:eproj_diss_n}, however, eigenprojection dissonance can resist $1/\sigma^2$ convergence for values of $\sigma$ considerably larger than $\rho$ 
(see the \verb|wine| and \verb|wdbc| data sets, in particular). 
Convergence onset remains positively correlated with $\rho$, but the strength of the dependence decreases with the
number of eigenvalues; eigenprojection convergence onset bandwidth increases with the number of eigenvalues. 
See Table~\ref{table:rhosVsEigdissonanceOnsetForLargeSigma}. Least-squares linear regression for convergence onset
with $\rho$ and eigenvalue number as independent variables yields coefficient values of $0.97$ and $0.62$
for these variables, respectively ($r^2 = 0.58$).

Table~\ref{table:rhosVsEigdissonanceForLargeSigma} shows that 
eigenprojection dissonance at $\sigma = 2^7$ remains correlated positively with data eccentricity when several eigenvalues are considered; however, the strength of this correlation decreases with the number of eigenvalues,
while the magnitude of eigenprojection dissonance increases with the number of eigenvalues.

We expect that correlation between $\rho$ and eigenprojection dissonance will remain close to $1$ for any number of eigenvalues at sufficiently high Gaussian bandwidths, as $1/\sigma^2$ convergence will have begun for all data sets at this point. 
As noted earlier in section~\ref{section:convergenceOnset}, the available numerical precision bounds the 
useable standardized bandwidths to those below $\sigma = 2^7$ or so.
We have not attempted an extended precision implementation that would enable reliable experimental values at bandwidths
higher than $2^7$.

\begin{table}[t!]
\caption{Convergence onset bandwidths (section~\ref{section:convergenceOnset}) track data eccentricity (Eq.~\ref{eq:rho_single_Gaussian_kernel}) closely.
Starred data set names are abbreviated (c.f., Table~\ref{table:dataSets}). Maximum reliably computed onset is $128$;
onset is shown as $256^+$ when convergence was not observed within the tested bandwidth range.
Least-squares linear slopes between $\log_2 \rho$ and $\log_2 \sigma^*_0$ (with coefficients of determination, $r^2$), 
for RKHS metric dissonance, norm of eigenvalue difference, eigenprojection dissonance (top eigenvalue only), respectively, are 
$0.86\ (r^2=0.96)$, $0.83\ (r^2=0.93)$, $0.88\ (r^2=0.92)$.} 
\begin{center}
\begin{tabular}{lrrrr}
\toprule
&&\multicolumn{3}{c}{Convergence onset, $\log_2 \sigma^*_0$}\\
\cmidrule{3-5}
Data set 	&$\rho$ 		&RKHS 	&eval 	&eigproj diss.\\
& 						&diss. 	&diff.	&(top ev only)\\
\hline
splice 				&1.15	&1	&2	&1\\	
tic-tac-toe 			&1.22	&1	&2	&0.5\\
wdbc 				&10.52	&8	&8	&8\\
diabetes 			&8.41	&8	&16	&8\\
wine 				&5.0		&4	&4	&4\\
balloon 				&14.83	&16	&16	&8\\
cpu 					&8.92	&8	&16	&4\\
boston 				&3.68	&4	&4	&2\\
stock 				&2.55	&2	&2	&2\\
cloud 				&4.67	&4	&8	&4\\
credit-g 			&10.34	&8	&8	&8\\
blood-transf-sc* 	&16.29	&16	&16	&8\\
monks-probs-2* 		&1.22	&1	&2	&1\\
steel-plates-fault 	&10.4	&8	&16	&4\\
kr-vs-kp 			&1.63	&2	&2	&0.25\\
qsar-biodeg 			&8.5		&8	&8 	&8\\
phoneme 				&2.57	&2	&2	&2\\
ozone-level-8hr 		&4.79	&4	&4	&4\\
hill-valley 			&41.42	&32	&64	&32\\
kc2 					&502.54	&256$^+$&256$^+$	&128\\
climate-mod-sc* 		&3.17	&4	&4	&2\\
spambase 			&106.69	&128	&128	&64\\
kc1 					&252.8	&256$^+$&256$^+$	&256$^+$\\
ilpd 				&33.4	&32	&64	&32\\
pc1 					&608.41	&256$^+$	&256$^+$	&128\\
pc3 					&1806.03	&256$^+$	&256$^+$	&256$^+$\\
one-hun-pl-mrg* 		&2.98	&4	&2	&2\\
one-hun-pl-shp*		&6.6		&8	&8	&4\\
one-hun-pl-txt*		&3.37	&4	&4	&4\\
banknote-auth* 		&3.87	&4	&4	&2\\
\bottomrule
\end{tabular}
\end{center}
\label{table:convergenceOnsetsVsRhos}
\end{table}

\begin{table}
\caption{Eigenprojection convergence onset (section~\ref{section:convergenceOnset}) is positively correlated with data eccentricity $\rho$
for multiple eigenvalues, but strength of dependence decreases with their number. 
Least-squares linear slopes, intercepts, $r^2$ for $\log_2 \rho$ vs.\ $\log_2$ eigenprojection convergence onset shown below.}
\begin{center}
\begin{tabular}{cccc}
\toprule
No.\ of eigenvalues &slope &intercept &$r^2$\\
\hline
1 &0.88 &-0.35 	&0.92\\
2 &0.93 &0.67 	&0.69\\
3 &0.87 &1.39 	&0.49\\
4 &0.85 &1.81 	&0.48\\
5 &0.83 &2.02 	&0.47\\
\bottomrule
\end{tabular}
\end{center}
\label{table:rhosVsEigdissonanceOnsetForLargeSigma}
\end{table}

\begin{table}
\caption{Large-$\sigma$ normalized eigenprojection dissonance (Eq.~\ref{eq:normalizedEigenprojectionDissonance}) increases approximately quadratically with data eccentricity,
$\rho$. 
Strength of dependence decreases with 
number of eigenvalues; magnitude of eigenprojection dissonance increases. 
Least-squares linear slopes, intercepts, $r^2$ for $\log_2 \rho$ vs.\ $\log_2$ eigenprojection dissonance at $\sigma=2^7$ shown below.}
\begin{center}
\begin{tabular}{cccc}
\toprule
No.\ of eigenvalues &slope &intercept &$r^2$\\
\hline
1 &2.05 &-19.74 &0.97\\
2 &2.13 &-18.45 &0.76\\
3 &1.99 &-16.18 &0.58\\
4 &1.98 &-15.38 &0.54\\
5 &1.94 &-14.94 &0.53\\
\bottomrule
\end{tabular}
\end{center}
\label{table:rhosVsEigdissonanceForLargeSigma}
\end{table}

\section{Conclusions}

We proved that, for a given data set, its RKHS representation corresponding to a Gaussian RBF kernel is asymptotically isometric as Gaussian bandwidth $\sigma \rightarrow \infty$ to an isotropic scaling of the original data representation in Euclidean space, which corresponds to a linear kernel. The residual is of order $O \left (\frac{\rho}{\sigma} \right )^2$, where $\rho$ is the ratio of maximum-to-median pairwise distance between examples in the original data set -- a measure of representation eccentricity. 

Our results suggest that constructions that rely on RKHS metric properties will, when applied to a Gaussian
kernel, produce results that are asymptotic to those of a linear kernel in the large-bandwidth limit.
Furthermore, $\rho$ provides a natural data-specific bandwidth scale, in the sense that near-linear behavior 
can be expected at bandwidths $\sigma \gg \rho$.
Recent results for Gaussian RBF CKA~\cite{AlvarezGaussianCKA2022} can be understood as a manifestation of this
deeper geometric picture. We provide a second illustration in the present paper, by showing that Gaussian RBF kernel PCA is asymptotically equivalent to classical (linear) PCA, again with an $O \left (\frac{\rho}{\sigma} \right )^2$ residual.
 
Our experimental validation confirms the robustness of our results over a variety of real data sets, and indicates that $\rho$ is a reliable and easily computable indicator of dataset-specific asymptotic behavior of Gaussian RBF kernel PCA, especially in the eigendirections associated with the largest eigenvalues. Therefore, the value of $\rho$ can be useful as a guide in selecting the value of the bandwidth hyperparameter $\sigma$ based on the specifics of the data sample. Given that finite-precision arithmetic limits the bandwidths that can be used in practice, our results may be less applicable to very high-eccentricity data sets.

\section{Future Work}

Our experimental results suggest that the $O(\rho/\sigma)^2$ convergence rate of Theorem~\ref{thm:GaussianPCAResidualDetails} is tight; however, they also leave open the possibility that the multiplicative constant $\eta$ in the bound can be lowered, and it would be of interest from a theoretical perspective to determine whether this is the case.
It would also be of interest to consider the case of repeated eigenvalues, and to characterize convergence of the eigenprojections associated with eigenvalues beyond the few largest, 
as data eccentricity alone appears to be less effective in that 
regard than for metric or eigenvalue convergence.

\small
\bibliography{IEEEabrv, KernelPCAReferences}
\normalsize

\end{document}